\documentclass[runningheads]{llncs}
 
\usepackage{subcaption}  
\usepackage{graphicx}  
\usepackage{booktabs}
\usepackage{array}
\usepackage{amsmath}
\usepackage[a4paper,margin=1in]{geometry}
\usepackage[T1]{fontenc}
\usepackage{graphicx}
\begin{document}

\title{Operationalizing Automated Essay Scoring: A Human-Aware Approach}

\author{Yenisel Plasencia-Calaña\inst{1}\orcidID{0000-0001-5173-0394}
}
%
 
%

\institute{Brigthlands Institute for a Smart Society, Smedestraat 2, 6411 CR Heerlen, The Netherlands 
\email {y.plasenciacalana@maastrichtuniversity.nl}\\
\url{https://www.biss-institute.com/}}
%

\maketitle              
\begin{abstract}
This paper explores the human-centric operationalization of Automated Essay Scoring (AES) systems, addressing aspects beyond accuracy. We compare various machine learning-based approaches with Large Language Models (LLMs) approaches, identifying their strengths, similarities and differences. The study investigates key dimensions such as bias, robustness, and explainability, considered important for human-aware operationalization of AES systems. Our study shows that ML-based AES models outperform LLMs in accuracy but struggle with explainability, whereas LLMs provide richer explanations. We also found that both approaches struggle with bias and robustness to edge scores. By analyzing these dimensions, the paper aims to identify challenges and trade-offs between different methods, contributing to more reliable and trustworthy AES methods.

\end{abstract}
\section{Introduction}

Automated Essay Scoring (AES) \cite{Mizumoto2023ExploringTP,Azhari2024OptimizationOA,Tang2024IncorporatingFL} is an interdisciplinary field at the intersection of Artificial Intelligence and Education, designed to support the essay grading process. These systems aim to overcome human limitations, including grading accuracy, human biases, and associated costs. Manual grading usually is time-consuming and slow \cite{RODRIGUES201430}, resulting in prolonged waiting times for students to receive their grades. Besides, there are high costs associated due to the need for professional and experienced evaluators. Other challenges include the risk of errors from inherent biases, due to fatigue from evaluating large volumes of essays, and inconsistencies in grading outcomes.

AES can assist educators in multiple ways, the most widely accepted application being formative assessment. In this context, AES provides students with immediate feedback on their writing, allowing them to learn about the errors they make, identify improvements, and refine their work before final submission. By offering constructive insights on grammar, structure, organization, and coherence among other dimensions, it facilitates learning and helps students to refine their writing skills over time.

Another important application is summative assessment, where AES supports educators in grading tests and assignments. By automating the evaluation process, AES reduces the time and effort required for large-scale grading, while ensuring efficiency and consistency. This could be particularly useful in large scale grading such as standardized tests and online education, where manually grading essays can be labor intensive. The goal is to use AES as a tool to assist educators, where the decision making and feedback still depend on the educator, and not as replacement of human judgment.

Over the years, numerous attempts have been made to develop AES systems for practical use. Notable examples include e-rater \cite{Attali2013AutomatedES}, designed to support Common Core Standards in both classroom settings and large-scale state assessments, and IntelliMetric \cite{intelliMetric}, developed for scoring open-ended essays. In recent years, AES systems have evolved due to advancements in text classification methods, particularly by leveraging progress in transformer models \cite{He2020DeBERTaDB} and Large Language Models (LLMs). Early AES systems, such as e-rater and IntelliMetric, relied on manually designed features, including grammar, coherence, and lexical diversity, to predict scores. More recently, deep learning models, especially transformer-based architectures like BERT, RoBERTa, and DeBERTa \cite{He2020DeBERTaDB}, have revolutionized NLP by accurately representing text semantics and context, improving accuracy in different text classification tasks including AES.

Current main approaches in the AES literature include models based on handcrafted features \cite{Tang2024IncorporatingFL}, transformers-based models \cite{Azhari2024OptimizationOA} such as DeBERTaV2 \cite{He2020DeBERTaDB} and DeBERTaV3 \cite{He2021DeBERTaV3ID}, tree-based classifiers\cite{Li2007McRankLT,Ke2017LightGBMAH}, and neural networks \cite{Uto2021ARO} such as Long Short-Term Memory (LSTM) \cite{Hochreiter1997LongSM}. More recently, we have witnessed the progress in Large Language Models (LLMs) such as GPT-4, which have introduced new paradigms for AES through zero-shot \cite{Mizumoto2023ExploringTP,Yavuz2024UtilizingLL} and few-shot learning \cite{Song2024AutomatedES}, eliminating the need for task-specific training or fine-tuning.

Most studies in the literature focus on optimizing the accuracy of AES systems. However, for AES systems to be adopted and used in a human-aware manner, accuracy is just one of several important dimensions. Other factors must be considered to develop a system that is practical and ethically sound. In this paper, we explore the challenges of operationalizing AES from a human-aware perspective for the main to groups of AES approaches in the literature. We state that there is a gap between the theoretical advances in AES and the requirements for operationalizing AES systems. We examine operationalization through key dimensions such as robustness, bias and explainability. For example, bias in AES models can disproportionately affect certain demographic groups, while lack of explainability can hinder trust and adoption in educational settings.
This work explores the human-aware operationalization of AES, emphasizing key dimensions beyond accuracy. We investigate:
\begin{itemize}
    \item Robustness: How different AES approaches handle stability of output scores, confidence in predictions, computational demand, and edge cases?
    \item Fairness: Whether AES models exhibit biases between demographic groups.
    \item Explainability: The extent to which AES systems can explain their scores and provide human-interpretable feedback.
\end{itemize}
To address these challenges, we pose the following research question:
What challenges and differences do the main type of AES approaches present in relation to human-aware operationalization dimensions such as robustness, bias, and explainability?
Through comparative analysis of ML-based and LLM-based AES methods, this paper highlights trade-offs between different techniques and provides insights into their suitability for practical use in educational settings.
\section{Methods}
We selected a set of methods that represent each of the current approaches in the literature to develop our analysis. They are the following:
\begin{itemize}
    \item Methods based on handcrafted features. We combine Word2vec, term frequency and inverse document frequency (TF-IDF), readability, text complexity, and grade level features. On top of these features we use either Histogram gradient boosting \cite{Li2007McRankLT} classification or regression, and LightGBM \cite{Ke2017LightGBMAH} regression.
    \item Method based on transformer features DeBERTa, either a finetuned DeBERTaV3 \cite{DBLP:conf/iclr/HeGC23} or DeBERTaV2 embeddings \cite{He2020DeBERTaDB} + Long Short-Term Memory networks (LSTM) \cite{Hochreiter1997LongSM} as classifier. We chose the DeBERTa model since it has set new standards in key Natural Language Processing (NLP) domains, including language comprehension, text generation, and question answering. DeBERTa leverages a self-attention mechanism to process sequential data, which allows the model to analyze all input tokens and create contextualized representations for each token based on its surrounding context. LSTM is a type of Recurrent Neural Network which allows to capture long term dependencies in the data. We specifically use Bidirectional LSTM (BiLSTM), which processes information both forward and backward, usually improving performance in NLP tasks.
    \item Methods based on LLMs. Recently, the progress on LLMs and the accuracy they show on different tasks, have made them attractive for different text recognition tasks, including AES \cite{Johnson2024ExaminingTR}. The LLM can be prompted with instructions, the essay to be scored and a rubric, and it provides a grade to the essay based on the rubric. Practitioners have also introduce new approaches to LLM use based on what is called agentic design patterns, where the LLM is enhanced either with data from which it can learn, tools, or processes that can help the LLM to generate higher quality answers. In AES, agentic design patterns such as few-shot learning and chain-of-thought have been recently introduced \cite{Lee2023ApplyingLL}. We analyze these approaches because they represent a shift from the traditional state-of-the-art in AES, which until now has relied solely on machine learning (ML) methods.
\end{itemize}
Once an AES model is trained, a key question emerges: what characteristics make it suitable for practical application? In this paper we analyze aspects that are important from a human-aware perspective. We focus on robust operating conditions, fairness, and explainability. Some studies in the literature of AES have focused on fairness \cite{Yang2024UnveilingTT}; however, they do not address other dimensions, or compare ML-based approaches with LLMs, or explore other fairness metrics that are more widely used in the machine learning field.

\subsubsection{Human-Aware Approach to Robust Operating Conditions.}
The operating conditions of an AI system encompass the factors and parameters that affect a prediction model's performance across different dimensions. In this section, we examine key dimensions of the main AES approaches, with a particular focus on their impact on the primary stakeholders in the AES process: the students being assessed and the educators responsible for grading.

\textbf{System Accuracy.} Although accuracy is not the primary focus of this paper, we include it to provide a more comprehensive view of the potential of AES systems. An accuracy metric should reflect the effectiveness of the system for the given application. Some of the frequently used metrics in supervised classification systems include classification accuracy, area under the receiver operating characteristics curve (AUROC), sensitivity, specificity, and the F1 score. However, AES has different characteristics compared to standard ML or supervised classification systems, since misclassifications or disagreements have different weights depending on the severity. Therefore, Cohen's Kappa is used in AES because it measures the level of agreement between human raters and the automated scoring system while accounting for agreement that could happen by chance. Unlike simple accuracy, Cohen's Kappa provides a more robust evaluation of the model’s reliability and consistency. High Cohen's Kappa values indicate that the system is consistently grading essays in a way that aligns with human judgment as given by training set labels. The more commonly used version is the Quadratic Weighted Kappa (QWK) because it penalizes larger disagreements (e.g., scoring an essay as a 1 instead of a 5 is worse than scoring it a 3 instead of a 4). A high Kappa implies high alignment with human raters, which is desirable to foster trust in the system. However, its worth noticing that even human raters often disagree when evaluating the same essays, which raises concerns about the objectivity of using human judgment as ground truth.

\textbf{Robustness to edge scores.} The ratio of instances in different classes (or score categories in the case of AES) can affect performance. In AES datasets, score distributions typically resemble a normal distribution, where most scores cluster around the mean, while very low and very high scores are more rare. This class imbalance poses a challenge for machine learning models, which tend to struggle with underrepresented categories. Mishandling these edge cases can have significant consequences: high-performing students can feel demotivated if their achievements are not properly recognized or in the extreme case can unfairly fail a test, while low-performing students may miss opportunities to learn and improve if their errors are not caught. To mitigate this issue, common strategies include oversampling the minority class, undersampling the majority class, or applying balanced class weights directly in the classifier. However, even when using these techniques, ensuring robust performance often still requires human oversight. In contrast, zero-shot LLM approaches for AES should be less affected by class imbalance, as they do not rely on training on labeled essay data for grading. We explore this topic further in the experiments section.

\textbf{Stability of results.} Settings such as learning rate, regularization, or randomization of some parameters influence the performance and stability of AES systems. This influence both the accuracy of predictions and the stability of the score when using the same model in different occasions to grade the same essay. For ML approaches, achieving stable scores can be easier by eliminating the randomization of certain parameters and setting a fixed randomization seed for them. In the case of LLMs, reducing the temperature parameter helps ensure more consistent outputs by minimizing variability. When the parameter is set to a low value (<0.5), the LLM provides stable scoring.

\textbf{Resource use.} Memory and computational power can constrain the size and complexity of the model, sometimes affecting accuracy if the model is too small or generating a high carbon footprint if the model is too large with major resource consumption. The different AES approaches vary in their computational requirements during training. Methods based on handcrafted features require fewer computing resources, whereas transformer-based models (e.g., DeBERTa v2 or v3), and neural networks require moderate size computing resources, while LLMs require major computing resources. Therefore, deciding on smaller AES models due to less resource consumption could be more attractive from an ethical perspective, while potentially compromising on accuracy. However, at inference time, computational requirements generally decrease for most approaches. The exception is LLMs utilizing chain-of-thought reasoning, which requires additional processing.

\textbf{Confidence Calibration.} Prediction models may need to provide reliable confidence scores alongside their predictions. When combined with the grade of the essay, these confidence scores become crucial for decision-making with human oversight. Human graders using an AES system for decision support can make more informed judgments about whether to assign a grade, based on the model's confidence in its prediction. Many machine learning approaches can provide a confidence score alongside their predictions, which may assist human graders in deciding whether to rely on the model’s output. In contrast, large language models (LLMs) typically do not offer a clear or calibrated measure of confidence, making it more difficult to assess the reliability of their responses.

\subsubsection{Bias and Fairness.}
We refer to fairness as ensuring consistent performance of the models in different demographic groups. By using fairness metrics, we can see where our models decision may cause disparate treatment to protected or sensitive groups. Some works from the literature already touch on the aspect of fairness of AES, such as \cite{Litman2021AFE}. In this work, the fairness of three models was evaluated based on standard metrics in the AES literature such as Overall Score Accuracy (OSA),Overall Score Difference (OSD), and Conditional Score Difference (CSD), which in general measure the variance between true scores and predicted score that can be explained by demographic attributes. However, when we look at fairness in the Machine Learning literature we find other approaches. For example, the metrics in the literature include demographic parity \cite{chouldechova2017fair}, equal oportunity \cite{hardt2016equality}, equalized odds  \cite{hardt2016equality}, predictive parity \cite{chouldechova2017fair} and treatment equality \cite{friedler2019fairness}. We can also analyze bias in the data directly by analyzing the predictive value of demographic attributes on the score.
For bias mitigation, some approaches look at how specific features influence bias, and in case of significant influence the feature can be removed. Other approaches to mitigate bias include balancing the training set for the affected demographics. An important aspect in bias mitigation is the awareness of how fairness and accuracy are often conflicting. Therefore, based on the application, a compromise between both should be carefully addressed.
An example of A fairness metric \cite{Yang2024UnveilingTT} from the AES literature is The Overall Score Accuracy (OSA), which measures whether AES scores are equally accurate across student groups compared to human scores. To compute this, the squared difference between human (\( H \)) and AES (\( S \)) scores is first calculated:

\[
(S - H)^2
\]

Fairness is evaluated by fitting a linear regression model where the squared error is the dependent variable and student demographics (e.g., gender) are the independent variables. The regression \( R^2 \) value serves as the OSA fairness metric, with statistical significance suggesting potential bias in the AES. A larger \( R^2 \) indicates that student group membership has a larger impact on score accuracy, meaning less fairness and more bias.
Anotehr metric, the Overall Score Difference (OSD), measures whether AES and human scores are consistently different across student groups. To maintain the sign of the difference, this calculation uses the absolute error rather than the squared error. \\
Other approaches used extensively in the machine learning literature rely on fairness metrics using the concepts of true positive rates and false positive rates. An example is the Equal Opportunity, which is satisfied by an AI system if it has equal true positive rates (TPR) across different demographic groups. It ensures that students who truly deserve a high score, are equally likely to receive one across different demographic groups. It is calculated as:
\begin{equation}
P(\hat{Y} = 1 \mid Y = 1, G = g_1) = P(\hat{Y} = 1 \mid Y = 1, G = g_2)
\end{equation}
where:
\begin{itemize}
    \item \( \hat{Y} \) is the predicted score (high or low),
    \item \( Y \) is the true score (ground truth),
    \item \( G \) represents the demographic group (e.g., male/female, native/non-native speakers).
\end{itemize}
Another example is the Equalized Odds metric. An AI system satisfies Equalized Odds if both the true positive rate (TPR) and false positive rate (FPR) are the same across demographic groups. True Positive Rate is the ratio of correctly identified positive cases to the total actual positive cases.
False Positive Rate is the ratio of incorrectly identified negative cases to the total actual negative cases. Therefore, it is expressed as:  
\begin{equation}
\text{TPR}_1 = \text{TPR}_2 \quad \text{and} \quad \text{FPR}_1 = \text{FPR}_2
\end{equation}
where \(\text{TPR}_1\) and \(\text{TPR}_2\) are the true positive rates for two demographic groups, and \(\text{FPR}_1\) and \(\text{FPR}_2\) are the false positive rates for those same groups. 
In practice, this means ensuring that both overscored and underscored students are treated fairly. Given the focus on fairness for both overscored and underscored groups, we will use the equalized odds metric in our experiments. Additionally, we extend equalized odds to a multiclass framework, allowing us to analyze bias across each score category. 
\subsubsection{Explainability.}
In this context, we define explainability as the system's ability to clarify how it arrives at its outcomes. Depending on the methods used, the level of explainability may vary. Simpler classifiers, such as decision trees, are generally more inherently explainable, while neural networks (especially deep ones) are often considered less explainable and are referred to as 'black-box' approaches. Despite the differences in inherent explainability, general explainability strategies can be applied to any classifier, often using feature importances to provide explanations at both the population and individual levels.
Explainability is crucial for any AI system, particularly in high-stakes applications where decisions significantly impact users, such as in health, education, or career-related decisions. It also plays a key role in fostering trust in the system. For both summative and formative assessments, explanations are essential as they provide insights into why a specific score was assigned. In formative assessments, these insights can guide students in improving unsatisfactory aspects of their responses. In the context of summative assesment, explainability can increase trust and facilitate the adoption of tools that support or enhance human grading.
Moreover, the relevance of explanations varies for the different stakeholders \cite{LANGER2021103473}. For essay evaluators, explanations should focus on how the rubrics are reflected in the essay to justify the grade. For students, the explanations should support their learning journey and the essay-writing process.
\section{Experiments}
The experiments aim to illustrate the differences among methods and their challenges in relation to various operationalization dimensions. We begin by comparing the accuracy of the different AES alternatives. Additionally, we conduct experiments on robustness to edge cases, fairness and explainability to analyze how each approach performs in these areas and assess their suitability for the essay scoring application.\\
We present the parameters used for each of the methods. For the Histogram Gradient Boosting classifier, we set the learning rate to 0.1, so the model makes gradual updates minimizing overfitting. The model is trained for a maximum of 100 iterations, the number of leaf nodes per tree is 30, while the maximum tree depth is set to 3. To prevent overfitting we set the minimum number of samples per leaf to 20. The same parameters were used for Histogram Gradient Boosting regression, and similar parameters were used for LightGBM: maximum tree depth=3 and number of leaves=30.\\
For the Long Short-Term Memory (LSTM) model, we used a two-layer architecture with 256 hidden units in the first layer and 128 hidden units in the second layer. To prevent overfitting, we applied a dropout rate of 0.5 and a recurrent dropout rate of 0.4. The input size was set to 768, corresponding to the embedding dimensions of the DeBERTa encoding. We used the ReLU activation function to add non-linearity. Additionally, we employed a bidirectional LSTM architecture to capture both forward and backward dependencies in the text. The model was trained using a batch size of 16 for 150 epochs, Adam Optimizer and Cross-entropy as loss function. For the finetuning of DeBERTaV3, we used a batch size of 32 for 40 epochs.\\

For the LLM experiments, we selected GPT-4o-mini. We chose GPT-4o-mini over GPT-4o due to the latter's higher computational cost and greater carbon footprint, while the performance difference was marginal when comparing in some of the database prompts. We used OpenAI's API service to run the model. We set a low temperature parameter at 0.2 to achieve higher response stability. When using an LLM, we compare two approaches: zero-shot, and few-shot with chain-of-thought (CoT) \cite{Cohn2024ACP} reasoning. For the prompting strategy we instructed the model to adopt a persona of an expert essay evaluator, then we provide the rubrics and we ask the LLM to use the rubrics as guidance for the evaluation. Finally, we asked the LLM to return the result in a format that is suitable for parsing the score. The prompt for the zero-shot experiment is as follows: "You are an expert evaluator of student essays, and your task is to score an essay based on the rubric: Holistic Rating for Source-Based Writing. After reading the essay, assign a holistic score based. Act as an impartial evaluator. Return the score in a structure with the following format: score = {score value}."

For the CoT combined with few-shot experiment, we used 2 example essays for each score. The prompt is modified to include: "Learn how the grading is performed by analyzing these examples:" and we complement the prompt providing the essays and their scores. 
In the case of the chain-of-thought experiment, we include the few-shot learning as inspired by \cite{Cohn2024ACP} and we added to each essay and score pair the statement: "Based on the rubric, the student earned a score of:". We enhance the instruction of few-shot learning with chain-of-thought by adding: "Let's think step by step.". The rubric, which can be found online \cite{persuade}, was also incorporated into the prompt.
\subsubsection{Data}
For the experiments, we use the PERSUADE 2.0 \cite{CROSSLEY2024100865} corpus. It comprises over 25,000 argumentative essays produced by 6th-12th grade students in the United States for 15 prompts on two writing tasks: independent and source-based writing. The PERSUADE 2.0 corpus provides detailed individual and demographic information for each writer. It has full texts, holistic score, word count, prompt, task, source texts, gender, grade level, english language learner status, race/ethnicity, economic status, and disability status. As part of the experimental protocol, we used the same training and test dataset division that is present in the dataset, ensuring results are comparable to previous methods from the literature. 
\subsubsection{Accuracy evaluation}
As performance metric we use Cohen's Quadratic Weighted Kappa which evaluates the level of agreement between the classification system and a human rater, while accounting for the possibility of agreement occurring by chance. Cohen's Kappa is particularly useful for ordinal data, where the categories have a meaningful order (e.g., rating scales from 1 to 6). Especially, the quadratic weighting penalizes larger disagreements more heavily than smaller ones. This is especially relevant when the scale is ordinal because misclassifying by a large margin is considered worse than a minor misclassification. 

Eq. \ref{eq:kappa} presents the Quadratic Weighted Kappa formula,
\begin{equation}
\kappa = 1 - \frac{\sum_{i,j} w_{ij} O_{ij}}{\sum_{i,j} w_{ij} E_{ij}}
    \label{eq:kappa}
\end{equation}
\noindent where:
\begin{itemize}
    \item \( \kappa \) is the Quadratic Weighted Kappa coefficient, indicating the level of agreement.
    \item \( O_{ij} \) represents the \textbf{observed agreement matrix}, where each element \( O_{ij} \) indicates the frequency of instances rated as category \( i \) by one rater and category \( j \) by the other.
    \item \( E_{ij} \) is the \textbf{expected agreement matrix}, representing the agreement expected by chance, calculated from the marginal distributions of ratings.
    \item \( w_{ij} \) is the \textbf{weight matrix}, assigning penalties to disagreements. The quadratic weighting penalizes larger differences more severely and is defined as:
    \[
    w_{ij} = \frac{(i - j)^2}{(N - 1)^2}
    \]
    \item \( i, j \) are the indices of the rating categories.
    \item \( N \) is the total number of rating categories.
\end{itemize}
\begin{table}[h!]
\centering
\begin{tabular}{|c|c|c|}
\hline
\textbf{Sl. No} & \textbf{Range of Quadratic Weighted Kappa} & \textbf{Concordance} \\
\hline
1 & Negative       & poor              \\
2 & 0.01–0.20      & slight            \\
3 & 0.21–0.40      & fair              \\
4 & 0.41–0.60      & moderate          \\
5 & 0.61–0.80      & substantial       \\
6 & 0.81–1         & almost perfect    \\
\hline
\end{tabular}
\caption{Interpretation of Quadratic Weighted Kappa Values}
\label{tab:quadratic_kappa}
\end{table}
Table \ref{tab:quadratic_kappa} shows the interpretation of the Quadratic weighted Kappa values. \\
\textbf{Accuracy Analysis}
Table \ref{tab:acc_comparison} presents the results of our experiments comparing the performance of different AES approaches. The results indicate that machine learning methods achieve higher agreement with human raters (as reflected in higher Kappa scores) and are therefore more accurate in grading essays, assuming human-labeled grades serve as a reliable ground truth. For these specific types of essays, the handcrafted feature-based approach performs the best, with both histogram gradient boosting regression and classification yielding the highest accuracy.
Regarding the different LLM strategies (zero-shot, and few shot + chain-of-thought (CoT)) we do not observe a clear systematic difference among them. However, previous studies have reported improvements when using few-shot or CoT prompting.

\begin{table}[h]
    \centering
    \small
    \renewcommand{\arraystretch}{1.2} 
    \begin{tabular}{|l|c|c|c|c|c|c|c|c|}
        \hline
        \textbf{Prompt} & \textbf{ZeroShot} & \textbf{CoT+FS} & \textbf{DeBV3} & \textbf{LSTM} & \textbf{HistGradC} & \textbf{HistGradR} & \textbf{LightGBM} \\
        \hline
        Phones and driving & 0.35 & 0.42 & 0.37 & 0.31 & 0.64 & 0.60 & 0.61 \\
        \hline
        Car-free cities & 0.52 & 0.51 & 0.60 & 0.54 & 0.69 & 0.63 & 0.65 \\
        \hline
        Summer projects & 0.44  & 0.48 & 0.58 & 0.43 & 0.60 & 0.64 & 0.65 \\
        \hline
        A Cowboy Who Rode Waves & 0.43 & 0.38 & 0.50 & 0.40 & 0.64 & 0.49 & 0.48 \\
        \hline
        Mandatory extracurricular & 0.49 & 0.60 & 0.64 & 0.60 & 0.69 & 0.66 & 0.65 \\
        \hline
        Exploring Venus & 0.58 & 0.53 & 0.70 & 0.59 & 0.75 & 0.72 & 0.73 \\
        \hline
        Facial action coding & 0.58 & 0.52 & 0.72 & 0.59 & 0.79 & 0.73 & 0.72 \\
        \hline
        The Face on Mars & 0.51 & 0.53 & 0.56 & 0.39 & 0.68 & 0.64 & 0.64 \\
        \hline
        Community service & 0.54 & 0.57 & 0.55 & 0.30 & 0.58 & 0.57 & 0.57 \\
        \hline
        Grades for extracurricular & 0.47  & 0.48 & 0.55 & 0.37 & 0.63 & 0.58 & 0.58 \\
        \hline
        Driverless cars & 0.38 & 0.40 & 0.53 & 0.33 & 0.65 & 0.59 & 0.57 \\
        \hline
        Cell phones at school & 0.51 & 0.44 & 0.57 & 0.30 & 0.62 & 0.60 & 0.60 \\
        \hline
        Distance learning & 0.48 & 0.60 & 0.71 & 0.68 & 0.77 & 0.74 & 0.74 \\
        \hline
        Seeking multiple opinions & 0.25 & 0.33 & 0.64 & 0.45 & 0.39 & 0.40 & 0.65 \\
        \hline
        Does the electoral college work? & 0.50 & 0.60 & 0.70 & 0.67 & 0.72 & 0.73 & 0.73 \\
        \hline
    \end{tabular}
    \caption{Comparison of Quadratic Kappa scores for different AES approaches}
    \label{tab:acc_comparison}
\end{table}

\textbf{Robustness to edge cases.} We use confusion matrices (see eq. \ref{eq:conf}) to show the discrepancies between true and predicted scores for edge cases such as very low or high scores.
\begin{equation}
C = \begin{pmatrix}
C_{11} & C_{12} & \cdots & C_{1K} \\
C_{21} & C_{22} & \cdots & C_{2K} \\
\vdots & \vdots & \ddots & \vdots \\
C_{K1} & C_{K2} & \cdots & C_{KK}
\end{pmatrix}
    \label{eq:conf}
\end{equation}

\begin{figure}[h!]
    \centering
    \begin{subfigure}[b]{0.40\textwidth}
        \centering
        \includegraphics[width=\linewidth]{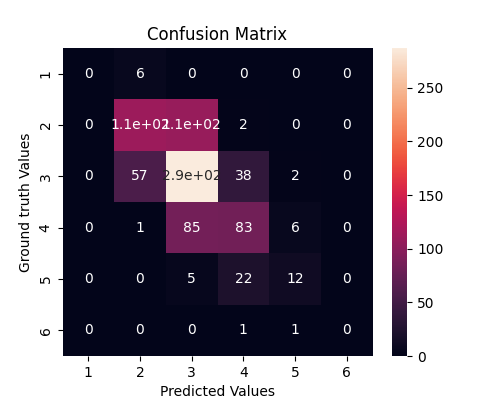}  
        \caption{Cell phones at school}
        \label{fig:cfcell}
    \end{subfigure} 
    \begin{subfigure}[b]{0.40\textwidth}
        \centering
        \includegraphics[width=\linewidth]{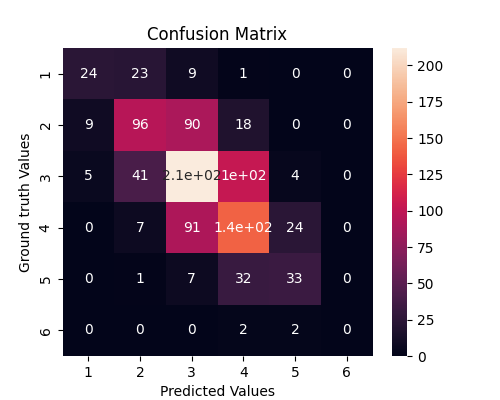}  
        \caption{Car-free cities}
        \label{fig:cfcar}
    \end{subfigure}
       \begin{subfigure}[b]{0.40\textwidth}
        \centering
        \includegraphics[width=\linewidth]{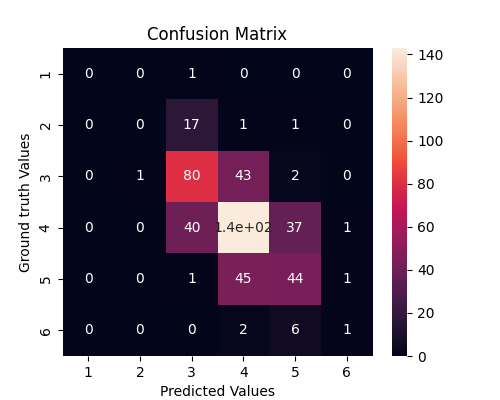}  
        \caption{Phones and driving}
        \label{fig:cfphones}
    \end{subfigure} 
    \begin{subfigure}[b]{0.40\textwidth}
        \centering
        \includegraphics[width=\linewidth]{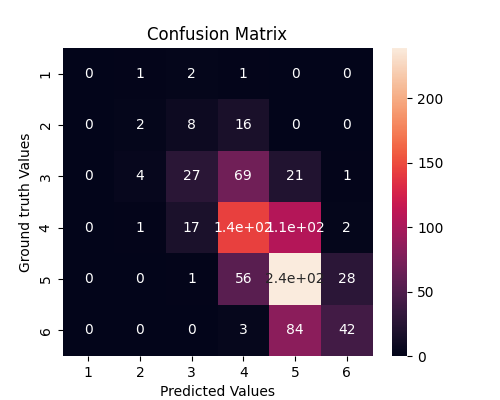}  
        \caption{Summer projects}
        \label{fig:cfsummer}
    \end{subfigure}
    \caption{Confusion matrices for four of the prompts in the database}
    \label{fig:confusion}
\end{figure}

\begin{figure}[h!]
    \centering
    \begin{subfigure}[b]{0.40\textwidth}
        \centering
        \includegraphics[width=\linewidth]{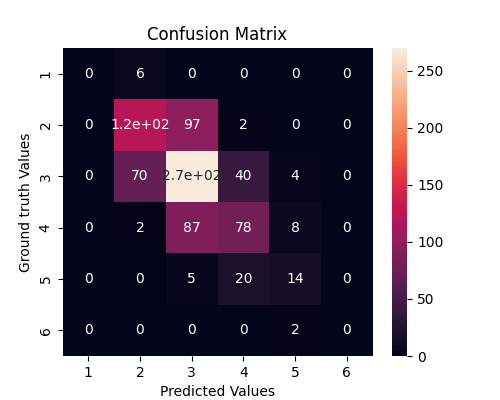}  
        \caption{Cell phones at school}
        \label{fig:cfcell}
    \end{subfigure} 
    \begin{subfigure}[b]{0.40\textwidth}
        \centering
        \includegraphics[width=\linewidth]{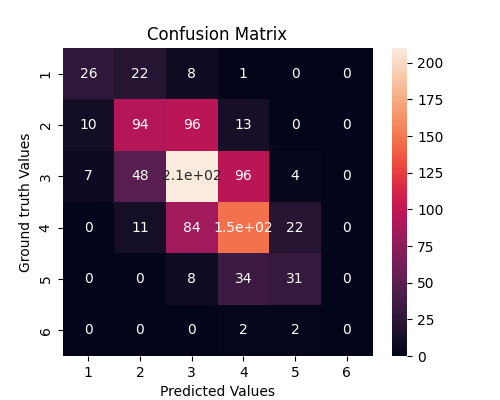}  
        \caption{Car-free cities}
        \label{fig:cfcar}
    \end{subfigure}
       \begin{subfigure}[b]{0.40\textwidth}
        \centering
        \includegraphics[width=\linewidth]{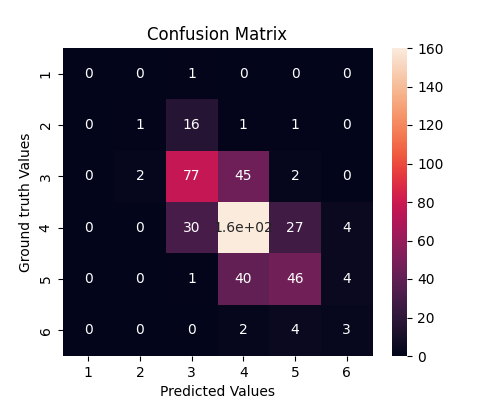}  
        \caption{Phones and driving}
        \label{fig:cfphones}
    \end{subfigure} 
    \begin{subfigure}[b]{0.40\textwidth}
        \centering
        \includegraphics[width=\linewidth]{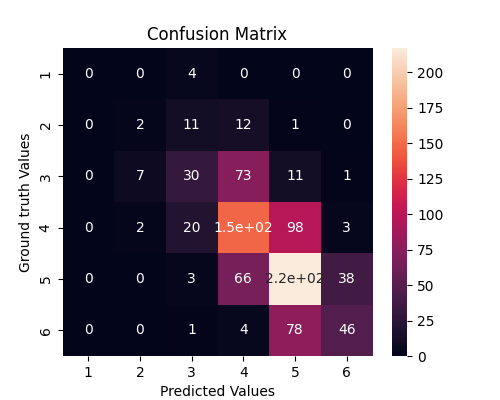}  
        \caption{Summer projects}
        \label{fig:cfsummer}
    \end{subfigure}
    \caption{Confusion matrices for four of the prompts in the database when balancing score labels}
    \label{fig:confusion_balanced}
\end{figure}

Where:
\begin{itemize}
    \item \( C_{ij} \) represents the number of instances of score \( i \) that were predicted as score \( j \).
    \item The diagonal elements \( C_{ii} \) represent the correct predictions for score \( i \).
    \item The off-diagonal elements represent the misclassifications.
\end{itemize}
For instance, examining the confusion matrices in Fig. \ref{fig:confusion} for the results of Histogram Gradient Boosting classifier (the most accurate method in our experiments) on some of the dataset prompts reveals that the model struggles to correctly assign scores at the extreme ends, such as 1, 2, 5, or 6. From Fig. \ref{fig:confusion_balanced} it can be seen that balancing class weights only mitigates the problem. The same issue occurs with the LLM, as it also struggles to assign true scores to edge cases. However, unlike the machine learning approach, there are no proven strategies to mitigate this problem.

\begin{table}[h!]
\centering
\begin{tabular}{|l|c|}
\hline
\textbf{Topic} & \textbf{Dem feat} \\ 
\hline
Phones and driving & 0.003 \\ 
\hline
Car-free cities & 0.11 \\ 
\hline
Summer projects & 0.27 \\ 
\hline
A Cowboy Who Rode the Waves & 0.13 \\ 
\hline
Mandatory extracurricular activities & 0.32 \\ 
\hline
Exploring Venus & 0.36 \\ 
\hline
Facial action coding system & 0.35 \\ 
\hline
The Face on Mars & 0.11 \\ 
\hline
Community service & 0.22 \\ 
\hline
Grades for extracurricular activities & 0.17 \\ 
\hline
Driverless cars & 0.11 \\ 
\hline
Cell phones at school & 0.15 \\ 
\hline
Distance learning & 0.57 \\ 
\hline
Seeking multiple opinions & 0.81 \\ 
\hline
Does the electoral college work & 0.02 \\ 
\hline
\end{tabular}
\caption{Demographic feature-based Quadratic Kappa for different prompts in the database}
\label{tab:dem_feat_comparison}
\end{table}

\begin{table}[h!]
\centering
\begin{tabular}{|l|c|c|c|c|c|c|}
\hline
\textbf{Bias GPT4omini and (HistGrad)} & \textbf{1} & \textbf{2} & \textbf{3} & \textbf{4} & \textbf{5} & \textbf{6} \\ 
\hline
Cell phones at school & 0.12 (0.001) & 0.03 (0.05) & 0.04 (0.02) & 0.03 (0.04) & 0 (0.05) & 0 (0) \\ 
A Cowboy Who Rode the Waves & 0.05 (0.06) & 0.12 (0.02) & 0.11 (0.07) & 0.01 (0.05) & 0 (0) & 0 (0) \\ 
Car-free cities & 0.06 (0.01) & 0.10 (0.04) & 0.02 (0.06) & 0.10 (0.03) & 0.03 (0.07) & 0 (0) \\ 
Community service & 0.20 (0) & 0.09 (0.08) & 0.12 (0.11) & 0.03 (0.03) & 0 (0.14) & 0 (0) \\ 
Distance learning & 0.33 (0) & 0.02 (0.01) & 0.07 (0.03) & 0.05 (0.02) & 0.01 (0.09) & 0 (0.05) \\ 
Does the electoral college work? & 0.06 (0.09) & 0.15 (0.09) & 0.11 (0.05) & 0.07 (0.04) & 0 (0.16) & 0 (0.25) \\ 
Phones and driving & 0 (0) & 0.11 (0.002) & 0.02 (0.01) & 0.03 (0.02) & 0.02 (0.02) & 0 (0.16) \\ 
Summer projects & 0 (0) & 0.03 (0.09) & 0.15 (0.02) & 0.10 (0.07) & 0.06 (0.06) & 0 (0.03) \\ 
Seeking multiple opinions & 0 (0)&	0.13 (0)&	0.08 (0.5)&	0.25 (0.58)	& 0 (0.62)&	0 (0.1) \\
\hline
\end{tabular}
\caption{Bias analysis based on the Equalized Odds metric for gender. The lower the value the better, where a value of 0 means no disparity. GPT-4o-mini and Histogram Gradient Boosting (result inside parenthesis) are compared for some of the dataset prompts}
\label{tab:bias_eq_odds}
\end{table}
\textbf{Bias Analysis.} For this analysis, we first applied an approach that attempts to predict the score based on demographic features using a Histogram Gradient Boosting classifier. If a high Kappa value is obtained, it may indicate that there is bias present in the scores, which the automatic grader could perpetuate. As it can be seen in Table \ref{tab:dem_feat_comparison}, some prompts show moderate to substantial agreement, with the "Seeking multiple opinions" prompt yielding a Quadratic Kappa of 0.81, indicating substantial to almost perfect agreement and therefore very high bias.

In Table \ref{tab:bias_eq_odds} we show the equalized odds for some of the database prompts. When analyzing the results from Table \ref{tab:bias_eq_odds}, we observe that, although gender could be expected to have lower bias compared to features like "Current language learner" for writing English essays, some bias is still present in the approaches. Interestingly, GPT-4o-mini sometimes exhibits higher bias than Histogram Gradient Boosting, even though the latter directly learns from the training data and can potentially perpetuate its biases. For the sake of space, we do not show all the results but we mention some of the most interesting insights.

For the "English learner status" feature, there is a larger bias compared to the gender, with values of 0.40 for some prompts. For the "Seeking multiple opinions" prompt, the disparate odds reach a value as high as 0.75 for score 4. A similarly high bias can be observed in Table \ref{tab:bias_eq_odds}, where scores for this prompt can be predicted with 81\% accuracy based solely on demographic features. Interestingly, GPT-4o-mini also shows substantial bias for scores 2, 3, and 4 for the same prompt, with values of 0.4, 0.3, and 0.12, respectively. However, these values are much lower than those shown by the trained model Histogram Gradient Boosting.

For the "Economically disadvantaged" feature, the equalized odds bias for Histogram Gradient Boosting never exceeds 0.23, except for the "Seeking multiple opinions" prompt, where bias reaches 0.41 for score 3, 0.58 for score 4, and 0.83 for score 5. GPT-4o-mini also shows some bias for score 4 (0.25), but this is much lower than the biases observed for Histogram Gradient Boosting.

A similar pattern is seen for the "Race/ethnicity" feature when comparing White race with other races and ethnicity. The disparity reaches 0.54 for score 4 and 0.91 for score 5 in the Histogram Gradient Boosting approach, while GPT-4o-mini shows bias primarily for scores 2 (0.28) and 3 (0.25), but these values are much lower than those observed for Histogram Gradient Boosting.

In edge cases, the bias is generally higher for Histogram Gradient Boosting compared to GPT-4o-mini, but GPT-4o-mini also demonstrates higher bias in similar extreme cases. 

\begin{figure}[h!]
    \centering
    \begin{subfigure}[b]{0.48\textwidth}
        \centering
        \includegraphics[width=\linewidth]{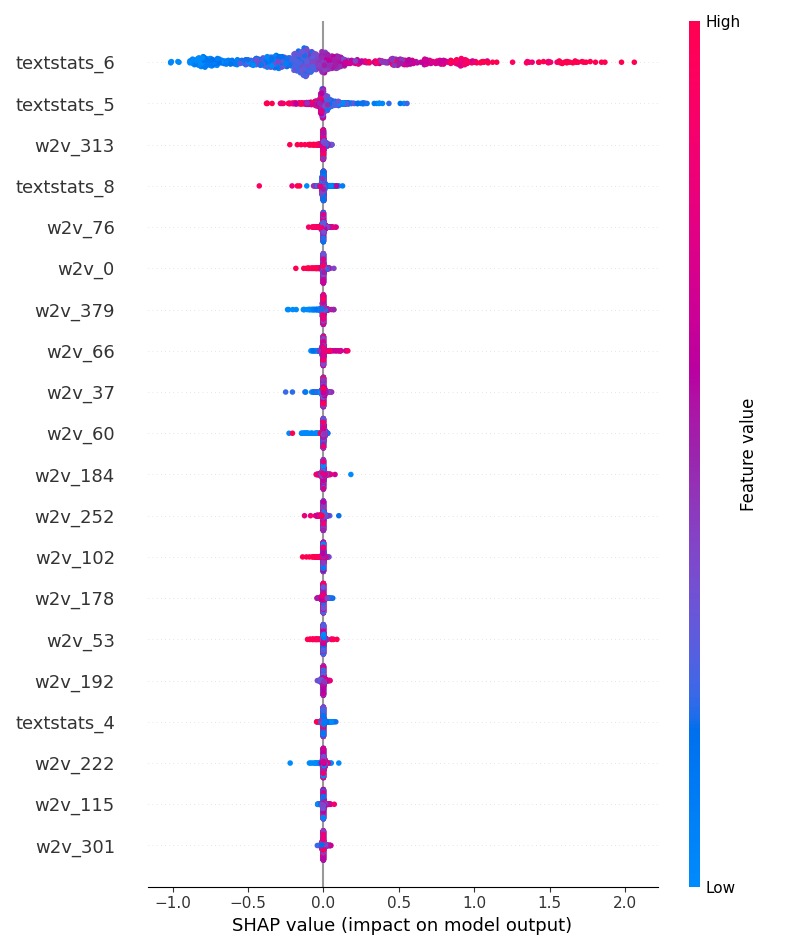}  
        \caption{Population-level explanations using SHAP us text stats and word2vec features.}
        \label{fig:shap_explanation}
    \end{subfigure}
    \hfill
    \begin{subfigure}[b]{0.48\textwidth}
        \centering
        \includegraphics[width=\linewidth]{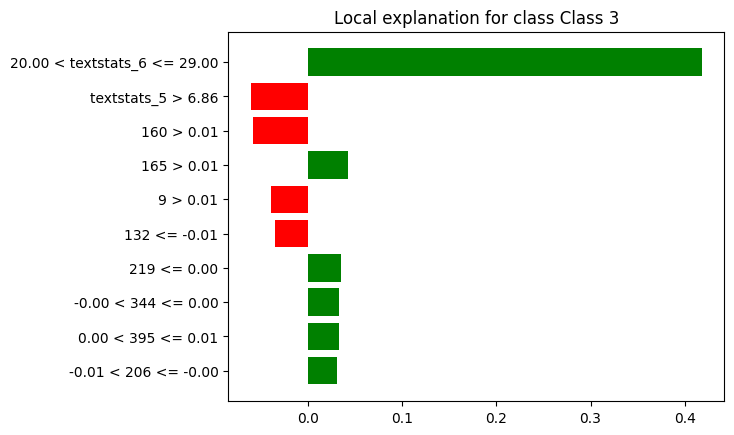}  
        \caption{Individual-level explanations by LIME using text stats and word2vec features.}
        \label{fig:lime_explanation}
    \end{subfigure}
    
    \caption{Comparison of SHAP and LIME explanations at both population and individual levels.}
    \label{fig:combined_explanations}
\end{figure}

ML-based methods could be perpetuating the bias in the training data of essays, while LLM-based methods could be perpetuating the bias from its training set. In conclusion, both approaches exhibit bias for various sensitive features.\\
\textbf{Explainability.}
Methods that rely on classifiers using handcrafted features, such as readability measures, are typically more explainable than those utilizing advanced features like RoBERTa embeddings. This is because higher or lower readability scores often have a direct relationship with the assigned grade, which can be observed through feature importances. For example, in Fig. \ref{fig:combined_explanations}, we show SHapley Additive exPlanations (SHAP) for feature importances at the population level for the "Cell phones at school" prompt. SHAP is a technique that quantifies the magnitude of each feature's contribution to the model’s output. From this, we can observe that text statistics features such as difficulty scores (e.g., Dale-Chall Readability Score, which gauges comprehension difficulty) and Gunning Fog Index (which estimates the years of formal education needed to understand the text) are highly influential, combined with some of the word2vec features. 
While this provides valuable insights into the model's decision-making process, there is still a gap between these feature-level explanations and more human-understandable justifications. In contrast, large language models (LLMs) can offer explanations that are more accessible to humans, detailing the reasoning process based on the rubrics and the essay itself. However, this does not necessarily explain how the model arrived at its decision based on its training, LLMs are considered "black boxes" in this respect, making it difficult to unveil their inner workings. Instead, LLMs provide reasoning for a given output and can even point to specific parts of the essay where the relevant issues are reflected.
In the methods section, we analyzed that the type of explanations that a student needs is different from the explanation that a professor needs, given their distinct goals in the AES task. While it is challenging to provide general explanations using standard feature importances and local LIME (Local Interpretable Model-agnostic Explanations), focusing these explanations to meet the specific needs of stakeholders presents an even greater challenge. In contrast, LLMs allow for more focused and tailored explanations, which can be achieved with relatively simple adjustments to the LLM’s prompt.

Below, we show an example of reasoning explanation by GPT-4o-mini for the same essay with true score = 3:\\

"Predicted score = 3.\\
The essay demonstrates developing mastery in presenting a point of view on the use of cell phones in schools. While it does articulate a perspective that advocates for allowing cell phone use during non-class times, several weaknesses detract from its effectiveness.

1. **Point of View**: The essay develops a point of view that is relatable, but it is somewhat vague and lacks depth in critical thinking. The arguments made about when cell phones should or should not be used are consistent, but they are not thoroughly developed or supported with strong examples or reasoning.

2. **Organization and Focus**: The organization of the essay is limited. There are some attempts to structure the ideas, but the progression of thoughts is not always clear, leading to lapses in coherence. For instance, the transitions between points could be improved to enhance clarity.

3. **Language Use**: The vocabulary is generally simple, and there is a noticeable lack of variety in sentence structure. While it communicates the message, the language lacks sophistication and expressiveness, which affects overall engagement.

4. **Mechanics**: There are several grammatical errors and awkward phrasings (e.g., "some of the school are not permitting the students to have") that can obscure meaning. These issues indicate that there is a need for improvement in mechanics and usage.

In summary, the essay presents a relevant argument, but the lack of depth in critical thinking, organization, and clarity in language use leads to a score that reflects developing mastery rather than a higher level of competence."\\
It can be seen that this explanation is more human understandable than the explanations we can obtain for the machine learning methods. The explanation can show where the problems are in the essay and how they determine the score given.


\section{Discussion and Conclusions}
When designing and implementing Automated Essay Scoring (AES) systems, factors beyond accuracy must be considered for human-aware operationalization. In this paper, we illustrate some of the challenges and differences of the main approaches found in the AES literature. To exemplify key concepts, we compared representative machine learning methods with LLM-based methods.

Our analysis reveals both similarities and differences. In terms of accuracy, some of the machine learning approaches outperform LLM-based methods, sometimes by a high margin across multiple prompts. However, we did not extensively optimize the ML methods, nor did we fine-tune the LLM, which could improve the results further.

Another key difference is the result stability. ML methods are generally able to produce consistent outcomes without significantly affecting accuracy. In contrast, LLMs can exhibit instability due to their inherent tendencies for creativity and response diversity. Setting the temperature parameter of the LLM to a low value (<0.5) overcomes this instability.

Regarding computational resource consumption, LLMs are among the most resource-intensive AI approaches during training. At inference time, the gap narrows unless agentic design patterns are incorporated into the prompt engineering process. Machine learning methods have from low to moderate resource consumption, making them a more attractive alternative to decrease carbon footprint and energy consumption. 

Both approaches also struggle with robustness to edge cases. For ML methods, class balancing strategies can partially mitigate this issue, but not completely. In the case of LLMs, we did not identify an effective strategy to address this limitation. However, fine-tuning the model may improve performance for this aspect.

In the bias dimension, both approaches exhibit biased outcomes. ML-based methods are known to replicate biases present in training data, while LLMs are expected to be more resilient to such biases (as not trained on the essay dataset). However, we found that LLMs still display bias in their outputs. Notably, for prompts where ML methods show high bias, LLM bias remains relatively low.

One area where LLMs excel is explainability. They can articulate their reasoning at varying levels of detail depending on the prompt. Nevertheless, the accuracy of such explanations has not been sufficiently investigated. In contrast, ML methods cannot achieve the same depth of interpretability regarding how an outcome is derived, even when leveraging widely used explainability techniques. This is a significant disadvantage, as explainability is crucial for fostering trust in AES systems and ensuring their effective use.

Our findings corroborate that evaluating accuracy alone is insufficient when designing and implementing human-aware AES systems. Even with high accuracy, widespread adoption requires addressing multiple additional factors, some of which we explored in this paper. Our analysis shows that the different type of AES methods still present bias, lack accurate explanations, and handle edge cases poorly. The adoption of AES in practice will be contingent on the development of systems that are not only accurate but also interpretable, fair, and robust.

\begin{credits}
\subsubsection{\ackname} We thank Prof. Rudolf Muller and Prof. Anna Wilbik for their insightful suggestions throughout the development of this work. 

\end{credits}

\bibliographystyle{plain}   
\bibliography{Operat_AES}   
\end{document}